\documentclass{article}
\usepackage{iclr2025_conference,times}

\usepackage{amsmath,amsfonts,bm}

\def\eqref#1{equation~\ref{#1}}

\def\1{\bm{1}}

\def\va{{\bm{a}}}

\def\vh{{\bm{h}}}

\def\vk{{\bm{k}}}

\def\vm{{\bm{m}}}

\def\vp{{\bm{p}}}

\def\vv{{\bm{v}}}

\def\vx{{\bm{x}}}

\def\mC{{\bm{C}}}
\def\mD{{\bm{D}}}

\def\mI{{\bm{I}}}

\def\mK{{\bm{K}}}

\def\mP{{\bm{P}}}
\def\mQ{{\bm{Q}}}
\def\mR{{\bm{R}}}
\def\mS{{\bm{S}}}

\def\mU{{\bm{U}}}
\def\mV{{\bm{V}}}
\def\mW{{\bm{W}}}

\DeclareMathAlphabet{\mathsfit}{\encodingdefault}{\sfdefault}{m}{sl}
\SetMathAlphabet{\mathsfit}{bold}{\encodingdefault}{\sfdefault}{bx}{n}

\usepackage{hyperref}
\usepackage{url}
\usepackage{graphicx}
\usepackage{subfig}
\usepackage{booktabs}
\usepackage{multirow}
\usepackage{amsmath}
\usepackage{amssymb}
\usepackage{bm}

\iclrfinalcopy

\title{Merging Methods for Multilingual Knowledge Editing for Large Language Models: An Empirical Odyssey}

\author{Kunil Lee$^{1,2}$, Ki-Young Shin$^{2}$, Jong-Hyeok Lee$^{1,3}$, Young-Joo Suh$^{4}$\thanks{Corresponding author: \texttt{yjsuh@postech.ac.kr}} \\
$^{1}$Department of Computer Science and Engineering, POSTECH, \quad
$^{2}$Designovel Co., Ltd. \\
$^{3}$LLSOLLU, \quad
$^{4}$Graduate School of Artificial Intelligence, POSTECH \\
\texttt{skywalker@postech.ac.kr}
}

\begin{document}

\maketitle

\begin{abstract}
Multilingual knowledge editing (MKE) remains challenging because language-specific edits interfere with one another, even when locate-then-edit methods work well in monolingual settings. This paper focuses on three issues: the effectiveness of vector merging methods for MKE, the extent to which Task Singular Vectors for Merging (TSVM) can reduce multilingual interference, and the influence of the weight scaling factor and rank compression ratio on performance. We evaluate six merging variants with two popular backbone large language models, two base knowledge editing methods, and 12 languages on the MzsRE benchmark under a large-scale batch-editing setting. Our results show that vector summation with shared covariance is the most reliable overall strategy, whereas simple summation without shared covariance performs poorly. TSVM improves performance in some settings, but its ability to mitigate multilingual interference is limited. We also find that performance is sensitive to both weight scale and rank ratio, with larger-than-default scaling and relatively low rank often yielding better results. These findings clarify the practical strengths and limits of current vector merging methods for MKE and provide guidance for future multilingual knowledge editing research.
\end{abstract}

\section{Introduction}
Modern large language models (LLMs) \citep{gpt, llama, qwen, gemini} contain billions to trillions of parameters, enabled by the scalability of the Transformer architecture \citep{attn}. Training such models requires substantial GPU compute and energy, and it produces considerable greenhouse gas emissions. To update knowledge stored in LLM parameters, efficient fine-tuning methods \citep{lora} have been developed, but they are still time-consuming \citep{mend}. Fortunately, knowledge editing (KE) provides a framework for modifying target knowledge while preserving unrelated knowledge at a much lower cost \citep{survey1}. In particular, \textit{locate-then-edit} KE methods can directly manipulate LLM parameters without additional learning \citep{rome, memit, alpha-edit}. However, most prior work has been conducted in English, and performance degradation has been observed in cross-lingual settings (e.g., edit in English $\rightarrow$ test in Chinese) \citep{cross}. Moreover, interference between languages has been reported in multilingual settings (e.g., edit and test in multiple languages simultaneously) \citep{lafn}. In particular, multilingual knowledge editing (MKE) performs substantially worse than monolingual KE given the same requests. Recently, a vector merging method called Task Singular Vectors for Merging (TSVM) was shown to effectively reduce task interference \citep{tsv}, which suggests a possible analogy to our problem.

In this paper, we explore the possibility of applying vector merging methods to multilingual knowledge editing (MKE). It is worth clarifying the difference between cross-lingual KE and MKE. In cross-lingual KE, evaluation is performed in languages that are not available at editing time. In MKE, by contrast, both editing and evaluation are performed in multiple languages simultaneously. We begin with the following research questions (RQs):
\begin{itemize}
    \item \textbf{RQ1:} How do vector merging methods perform in MKE?
    \item \textbf{RQ2:} Can TSVM effectively mitigate multilingual interference?
    \item \textbf{RQ3:} How do factors such as weight scale and rank ratio affect performance?
\end{itemize}

Although the analogy is promising, vector merging methods originate from a different problem: combining separately fine-tuned weights into a single weight update. We therefore design a systematic study of several merging methods while controlling for backbone type, base KE method, shared covariance, weight scale, and rank compression ratio. Our contributions are as follows:
\begin{itemize}
\item To the best of our knowledge, this is the first systematic study of locate-then-edit MKE with mass editing in 12 languages in parallel (batch size $=700\times12$). Previous work \citep{lafn, mzsre} used the same dataset \citep{mzsre}, but considered only one editing request across 12 languages at a time (batch size $=12$).
\item We evaluate six merging methods with two backbones and two base KE methods, and observe that vector summation with shared covariance achieves the strongest overall performance. We also show that TSVM can reduce interference under limited conditions, but in general none of the tested merging methods effectively closes the gap between MKE and monolingual KE.
\item To the best of our knowledge, our work is the first to analyze the weight scaling factor in MKE. Surprisingly, we find that the optimal scaling factor can exceed the default value of $1.0$.
\item We also investigate the effect of the rank compression ratio on TSVM performance and find that relatively low rank often leads to better results.
\end{itemize}

\section{Related Work}

\subsection{Knowledge Editing for Large Language Models}
Knowledge editing (KE) aims to update factual associations in large language models without retraining the entire model. Compared with conventional fine-tuning methods such as LoRA \citep{lora}, KE seeks to modify targeted knowledge while preserving unrelated behavior at substantially lower cost. Existing KE methods span several paradigms, including early hyper-network-based approaches such as KnowledgeEditor \citep{knowledge-editor}, learned editors such as MEND \citep{mend}, semi-parametric methods such as SERAC \citep{serac}, and in-context editing methods such as IKE \citep{ike}. Among these lines of work, our study is most closely related to direct parameter-editing approaches.

More specifically, locate-then-edit methods are motivated by the view that factual associations are stored in transformer feed-forward layers as key--value memories \citep{key-value}. ROME \citep{rome} first showed that factual knowledge can be localized and modified through closed-form updates to specific layers. MEMIT \citep{memit} extended this idea to the mass-editing setting, PMET \citep{pmet} improved editing precision, and AlphaEdit \citep{alpha-edit} further refined the approach by constraining updates with a null-space projection. Complementary work has also examined knowledge localization at the neuron level: Knowledge Neurons \citep{knowledge-neurons} identified neurons associated with factual recall and demonstrated small-scale factual editing through neuron intervention. Although these methods achieve strong results in monolingual settings, multilingual knowledge editing remains relatively underexplored. Prior work has shown that cross-lingual KE degrades when editing and evaluation are performed in different languages \citep{cross}, and more recent work has extended this challenge to cross-lingual multi-hop settings \citep{crolin-mquake}. More directly related to our problem, MzsRE \citep{mzsre} provides a multilingual benchmark covering 12 languages, and LU-LAFNs \citep{lafn} reduces interference by identifying language-agnostic factual neurons. In contrast, we focus on mass multilingual editing and ask whether vector merging can mitigate interference among language-specific edits.

\subsection{Task Vectors and Model Merging}
Model merging studies how to combine parameter updates from multiple tasks or fine-tuned models into a single model without retraining from scratch. The simplest strategies sum or average parameter differences, but these approaches often suffer from task interference when the underlying updates are not well aligned. This issue is closely related to our setting, where language-specific editing vectors may contain both shared information and conflicting components.

Representative work in this area includes Model Soups, which showed that averaging weights from multiple fine-tuned models can improve robustness when the solutions lie in a shared basin \citep{model-soups}; Task Arithmetic, which demonstrated that task vectors extracted from fine-tuned models can be composed through simple algebraic operations \citep{task-arithmetic}; and TIES-Merging, which explicitly addresses redundant updates and sign conflicts across task vectors \citep{ties-merging}. Our work is most directly inspired by Task Singular Vectors for Merging (TSVM) \citep{tsv}, which introduces a low-rank view of task updates and shows that singular-vector-based compression can reduce interference during model merging. Although TSVM was originally proposed for merging task-specific model updates rather than multilingual knowledge edits, its underlying motivation is highly relevant to MKE. This connection motivates our study of summation-, averaging-, and TSVM-based rules for merging language-wise editing vectors. At the same time, our setting differs from standard model merging: rather than combining independently fine-tuned task models, we merge editing vectors produced by locate-then-edit KE methods. Our work therefore brings together two previously separate lines of research, namely direct knowledge editing for LLMs and interference-aware vector merging.

\section{Preliminaries}
\subsection{Large Language Models}
Structurally, a Large Language Model (LLM) is an architecture built from stacked Transformer \citep{attn} blocks. Within each block, two distinct modules operate in tandem. First, an attention layer transforms contextual input into short-term memory. This is paired with a position-wise feedforward network (FFN) that selects \textit{active slots} from massive vector stores, effectively functioning as the model's long-term memory. Finally, all outputs from these modules are summed with the previous layer's activations in a residual fashion. More specifically, the output activation of $\vx$ at layer $l$ within the model, denoted as $\vh^l$, can be defined as:
\begin{equation}
         \vh^l= \vh^{l-1} + \va^l + \vm^l,
\end{equation}
\begin{equation}
         \quad\vm^l= \mW_{\text{out}}^l \,\sigma(\mW_{\text{in}}^l \, \gamma(\vh^{l-1}+\va^l)\,),\label{eq:ffn}
\end{equation}
where $\va^l$ and $\vm^l$ denote the outputs of the attention layer and the FFN layer, respectively; $\mW_{\text{in}}^l$ and $\mW_{\text{out}}^l$ denote the weight matrices of the FFN layers; $\sigma$ denotes the non-linear activation function; and $\gamma$ denotes the layer normalization. Here we follow expressions defined in \citep{rome, alpha-edit}.

\subsection{Linear Associative Memory}
A Linear Associative Memory (LAM) receives a key $\vk$ and recalls its corresponding value $\vv$ as follows:
\begin{equation}
         \vv = \mW\vk, \forall (\vk, \vv) \in \mS,
\end{equation}
where $\mW$ is the weight matrix of the LAM, designed to memorize $(\vk, \vv)$ associations within a finite set $\mS$. Under this framework, a FFN layer from \eqref{eq:ffn} can be interpreted as follows \citep{key-value}:
\begin{equation}
    \underbrace{\vm^l}_{\let\scriptstyle\textstyle
    \substack{\vv}}= \mW_{\text{out}}^l \,\underbrace{\sigma(\mW_{\text{in}}^l \, \gamma(\vh^{l-1}+\va^l)\,)}_{\let\scriptstyle\textstyle
    \substack{\vk}},\label{eq:lam}
\end{equation}
where the down-projection matrix $\mW_{\text{out}}^l$ serves as the LAM weight matrix.

\subsection{Locate-then-edit Methods}
In knowledge editing(KE), we study \textit{knowledge} (like facts) of the form (subject $s$, relation $r$, object $o$). For example, in the sentence ``The president of the U.S. is Joe Biden,'' the subject is ``the United States,'' the relation is ``the president,'' and the object is ``Joe Biden.'' An editing request is also a triple $(s, r, o \rightarrow o')$, where $o'$ is the new object we want to modify; for example, (``the United States'', ``the president'', ``Joe Biden'' $\rightarrow$ ``Donald Trump'').

The \textit{locate-then-edit} framework assumes critical information used to predict object $o$ given subject $s$ and its relation $r$, is stored within the key-value structure of the FFNs defined in the \eqref{eq:lam}. Like an standard LAM, the FFNs recall facts from certain input (such as a prompt about a subject and a type of relation) and contribute significantly to predict the next tokens associated with the correct object.

Under this framework, the overall KE process is divided into two steps: 1) identifying the specific Transformer layers responsible for storing knowledge, and 2) applying a KE algorithm to these layers sequentially in bottom-to-top order. The identification of these layers is conducted via causal tracing techniques \citep{rome, memit}. Once the identification have been established for a specific model architecture, further recomputation is generally unnecessary; consequently, subsequent research typically adopts these standardized layer ranges for editing tasks. For each target layer, the KE algorithm computes a matrix perturbation $\bm{\Delta}$ based on editing requests for the weight matrix $\mW_{\text{out}}$ defined in \eqref{eq:lam} (We omit the layer index $l$ in the rest of this paper for brevity). The resulting updated weight matrix, $\mW'_{\text{out}} = \mW_{\text{out}} + \bm{\Delta}$, is intended to encode the newly modified knowledge.

Formally, suppose we have $n$ editing requests and a set of corresponding keys and values for newly updated knowledge, $\mS_{1} = \{(\vk_{i}, \vv_{i}) | 0 < i \leq n \}$, and concatenate these keys and values into matrices, which are denoted as:
\begin{equation}
    \mK_{req}=\left[\vk_1\left|\vk_2\right| \ldots \mid \vk_n\right]\in \mathbb{R}^{h \times n},\quad \mV_{req}=\left[\vv_1\left|\vv_2\right| \ldots \mid \vv_n\right]\in \mathbb{R}^{d \times n};
\end{equation}
and similarly, $\mK_{const}$ and $\mV_{const}$ denote keys and values for other knowledge should be preserved. Then we have:
\begin{equation}
    \bm{\Delta} =  \mathop{\mathrm{arg\,min}}_{\bm{\tilde{\Delta}}}(\left\|({\mW_{out}}+\bm{\tilde{\Delta}}) \mK_{req}-\mV_{req}\right\|^2
    +\left\|({\mW_{out}}+\bm{\tilde{\Delta}}) \mK_{const}-\mV_{const}\right\|^2). \label{eq:memit-loss}
\end{equation}
We can obtain closed-form solution for this problem by applying normal equation \citep{linalg, memit}:
\begin{equation}
\bm{\Delta}_\textbf{MEMIT}=\mR \mK_{req}^T\left(\lambda\mC_{const}+\mC_{req}\right)^{-1},\label{eq:memit}
\end{equation}
where the error term $\mR=\left(\mV_{req}-{\mW} \mK_{req}\right)$; $\lambda$ is a hyperparameter; and covariance matrices $\mC_{const}=\mK_{const} \mK_{const}^T$ and $\mC_{req}=\mK_{req} \mK_{req}^T$. This method is often called \textbf{MEMIT}. If we apply null-space constrain in further \citep{alpha-edit}, we have \textbf{AlphaEdit} method as follows:
\begin{equation}
\bm{\Delta}_\textbf{AlphaEdit}=\mR \mK_{req}^T\mP\left(\lambda\mI+\mC_{req}\mP\right)^{-1},\label{eq:alpha-eidt}
\end{equation}
where $\mP$ is the null-space projection matrix (null-space of $\mC_{const}$); $\lambda$ is a hyperparameter. In the error term $\mR$, $\mV_{req}$ means ideal values that a perfectly modified model can recall for keys $\mK_{req}$, is usually calculated with approximation by ``network hooking'' methods \citep{rome, memit}.

In this paper, we only consider one-step batch editing scenario, which means applying KE algorithms at once with all editing requests. Sequential editing problem is out-of-scope of this work.

\section{Merging Methods for Multilingual Knowledge Editing}
\subsection{Problem Definition}
In a multilingual knowledge editing (MKE) scenario, suppose that we have $n$ editing requests and each request was written in $m$ languages, we can calculate $\Delta$ using \eqref{eq:memit} or \eqref{eq:alpha-eidt} for each language and apply it to the original model as follows:
\begin{equation}
\mW^{new} = \mW^{old} + \alpha \bm{\Delta}^{merged},\label{eq:update}
\end{equation}
\begin{equation}
\bm{\Delta}^{merged} = f(\bm{\Delta}^1, \bm{\Delta}^2, \ldots, \bm{\Delta}^m),\label{eq:merge-func}
\end{equation}
where $f(\cdot), \underbrace{\mathbb{R}^{d \times h}\times\ldots\mathbb{R}^{d \times h}}_{\let\scriptstyle\textstyle \substack{m}}\Rightarrow\mathbb{R}^{d \times h}$ denotes the merging function and $\alpha$ denotes the scaling factor. For each $\bm{\Delta}^i$ of i-th language, we call it knowledge editing vector for i-th language.

\subsection{Merging Functions}
In this paper, we study the following merging functions for \eqref{eq:merge-func}. To calculate covariance matrix in the \eqref{eq:memit} or \eqref{eq:alpha-eidt} for each language, we can either 1) gather covariance only in that language: $\mC_{req}^i = \mK_{req}^i(\mK_{req}^i)^T$; or 2) gather across all languages: $\mC_{req}^i = \sum_{j=1}^m\mK_{req}^j(\mK_{req}^j)^T$. We first introduce merging methods based on $\bm{\Delta}^i$s calculated under the former case as follows:
\begin{itemize}
\item	\textbf{Sum}:
\begin{equation}
\bm{\Delta}^{merged} = \sum^m_{i=1}\bm{\Delta}^i.
\end{equation}
\item	\textbf{Mean}:
\begin{equation}
\bm{\Delta}^{merged} = \frac{1}{m}\sum^m_{i=1}\bm{\Delta}^i.
\end{equation}
\item	\textbf{TSVM}\citep{tsv}: first, we decompose each $\bm{\Delta}^i$ into $\mU^i$, $\bm{\Sigma}^i$, $\mV^i$ using Singular Value Decomposition (SVD), then select top-k highest singular values and their corresponding column and row vectors from $\mU^i$ and $\mV^i$:
\begin{equation}
\tilde{\mU}^i = \mU^i\left[:,1:k\right], \tilde{\bm{\Sigma}}^i = \bm{\Sigma}^i\left[:k\right],\tilde{\mV}^i = \mV^i\left[1:k\right].
\end{equation}
The value $k$ is determined as follows:
\begin{equation}
k = \lfloor r \cdot d \rfloor,\label{eq:factor-r}
\end{equation}
where the factor $r$ is a hyperparameter. It reflects the degree of rank compression: if $r$ is equal to 1, that means no compression; if $r$ is close to zero, the $\bm{\Delta}$ is highly compressed into a low-rank matrix. Subsequently, we concatenate resulted components across all languages as follows:
\begin{equation}
\begin{split}
\mU^{concat} = \left[\tilde{\mU}^1 \mid \tilde{\mU}^2 \mid \ldots \mid \tilde{\mU}^m\right],\\
\bm{\Sigma}^{concat} =
\begin{bmatrix}
\tilde{\bm{\Sigma}}^1 & \bm{0} & \ldots & \bm{0} \\
\bm{0} & \tilde{\bm{\Sigma}}^2 & \ldots & \bm{0} \\
\ldots & \ldots & \ldots & \ldots \\
\bm{0} & \bm{0} & \ldots & \tilde{\bm{\Sigma}}^m
\end{bmatrix},\\
\mV^{concat} = \left[\tilde{\mV}^1 \mid \tilde{\mV}^2 \mid \ldots \mid \tilde{\mV}^m\right]^T.
\end{split}
\end{equation}
To make $\mU^{concat}$ orthogonal, we decompose $\mU^{concat}$ to $\mP$, $\mD$, $\mQ$, by applying SVD, and set $\mU^{merged} = \mP\mQ$ (similarly for $\mV^{concat}$). Finally, we have:
\begin{equation}
\bm{\Delta}^{merged} = \mU^{merged}\bm{\Sigma}^{concat}\mV^{merged}.
\end{equation}
\end{itemize}
In addition, we introduce merging methods based on $\bm{\Delta}^i$s calculated under the latter case (gather covariance across all languages), and name these methods with suffix ``-Cov'' to distinguish them with former ones.
\begin{itemize}
\item	\textbf{Sum-Cov}: same as Sum, but with $\bm{\Delta}^i$s calculated by shared covariance. This is similar to the method used in the previous work \citep{lafn} that adapted MEMIT to MKE. However, they apply the editing algorithm with one request (in multiple languages) at a time; on the contrary we apply the editing algorithm with massive requests (in multiple languages) simultaneously in one batch.
\item \textbf{Mean-Cov} and \textbf{TSVM-Cov}: same as Mean and TSVM, but with $\bm{\Delta}^i$s calculated by shared covariance.
\end{itemize}

\section{Experiments}
\subsection{Experimental Settings}
\subsubsection{Metrics}
We follow \citep{mend, rome, memit, alpha-edit}, use \textbf{accuracy} on four metrics: \textbf{Efficacy}, \textbf{Generalization}, \textbf{Specificity}, and \textbf{Portability}\citep{mzsre, lafn}. The accuracy is defined as follows:
\begin{equation}
    Acc(\vp, \va, P_{\theta}(t \mid \cdot)) = \mathbb{E}_{o_{i} \in \mathrm{tokenize}(\va)}\{o_{i} = \mathop{\mathrm{arg\,max}}_{t_{i}} P_{\theta}(t_{i} \mid \vp + t_{\leq i})\},
\end{equation}
where $P_{\theta}(t \mid \cdot)$ denotes an auto-regressive language model; $\vp$ denotes the input prompt; $\va$ denotes the anticipated answer given the input prompt.

\begin{itemize}
    \item \textbf{Efficacy} is the expectation of accuracy on editing requests:
\begin{equation}
    Acc_{eff} = \mathbb{E}_{\vp, \va \sim \mD_{req}}\{Acc(\vp, \va, P_{\theta}(t \mid \cdot))\}.
\end{equation}

    \item \textbf{Generalization} is the accuracy on paraphrases of the edit requests:
\begin{equation}
    Acc_{gen} = \mathbb{E}_{\vp, \va \sim \mD_{req}}\{\mathbb{E}_{\vp' \sim \mathrm{rephrase}(\vp)}\{Acc(\vp', \va, P_{\theta}(t \mid \cdot))\}\}.
\end{equation}

    \item \textbf{Specificity} is the accuracy on questions that are unrelated to editing requests:
\begin{equation}
    Acc_{spe} = \mathbb{E}_{\vp, \va \sim \mD_{req}}\{\mathbb{E}_{\vp' \sim \mathrm{unrelated}(\vp)}\{Acc(\vp', \va, P_{\theta}(t \mid \cdot))\}\}.
\end{equation}
    \item \textbf{Portability} is the accuracy on questions about one-hop reasoning of editing requests:
\begin{equation}
    Acc_{por} = \mathbb{E}_{\vp, \va \sim \mD_{req}}\{\mathbb{E}_{\vp' \va' \sim \mathrm{onehop}(\vp, \va)}\{Acc(\vp', \va', P_{\theta}(t \mid \cdot))\}\}.
\end{equation}
\end{itemize}
On top of these four accuracies, we report \textbf{averaged accuracy}, which is the arithmetic average of these four accuracies, as a unified metric in our experiments.

\subsubsection{Dataset}
We use the test split of MzsRE \citep{mzsre} in our experiments. It contains 700 samples, each includes one editing request, one paraphrase of editing request, one unrelated question and answer, one one-hop altered question and answer of editing request, which are written in 12 languages: English(en), Chinese(zh), Czech(cz), Vietnamese(vi), Turkish(tr), French(fr), Spanish(es), German(de), Russian(ru), Dutch(du), Portuguese(pt), Thai(th).

\subsubsection{Backbones \& Base Methods}
We use Llama3.1-8B-Instruct\citep{llama} and Qwen2.5-7B-Instruct\citep{qwen} as our backbones. Follow related work, we target critical layers [4,5,6,7,8] for editing, for both backbones.

For each backbone, we conduct our merging methods to two base methods: MEMIT\citep{memit} and AlphaEdit\citep{alpha-edit}. We choose MEMIT and its successor AlphaEdit because they are the most representative work under the locate-then-edit framework.

\subsubsection{Monolingual Editing}
We also evaluate monolingual editing performance for each language. It's done by applying editing algorithms with only one language. We compare this results with multilingual counterparts to analysis interference occurred in multilingual setting. They are named as ``Mono'' in the result tables.

\subsubsection{Other Details}
We use several NVIDIA A100 PCIe 80GB GPUs for our experiments and use pytorch \citep{pytorch} as development framework. Calculations of $\bm{\Delta}$, merging results are performed with \textit{float32} precision, estimating value matrix $\mV_{req}$ and evaluation is performed with \textit{bfloat16} precision.

\subsection{Experimental Results}
\subsubsection{Comparison of Averaged Accuracies of Merging Methods on MzsRE (\textbf{RQ1, RQ2})}
We have following observations from Table~\ref{tab1} and Table~\ref{tab2}.
\begin{itemize}
    \item \textbf{Observation 1: Sum-Cov achieves superior performance than other methods in most cases.} As contrast, simply summation without sharing covariance (Sum) records almost zero accuracy. Mean, which is the squeeze version of Sum, records improved performances. However, Mean-Cov performs much worse than Sum-Cov.
    \item \textbf{Observation 2: TSVM can significantly improve overall performances, while TSVM-Cov failed to achieve higher scores than Sum-Cov in most cases.} However, as shown in Table~\ref{tab2}, both TSVM and TSVM-Cov achieves higher accuracies than Sum-Cov, with Qwen backbone and AlphaEdit algorithm.
    \item \textbf{Observation 3: There are significantly large gap of performance between monolingual editing and multilingual ones.} Similarly, previous work \citep{lafn} reports the performance gap between MKE and monolingual editing with experimental settings that are different with ours. They conduct editing and evaluation with one request at a time while we conduct experiments with all requests at once. We further investigate the possibility of mitigating multilingual interference with merging methods, but unfortunately none of them effectively achieve the goal.
\end{itemize}

\begin{table}[t]
\caption{Comparison of merging methods on MzsRE dataset with backbone Llama3.1-8B. Averaged accuracies are reported, unit is percentage. The rank ratio $r$ for TSVM and TSVM-Cov are 0.125, 0.292 for MEMIT, 0.125, 0.063 for AlphaEdit, respectively. Weight scale is set to 1.0.}
\label{tab1}
\begin{center}
\resizebox{\textwidth}{!}{%
\begin{tabular}{c|cccccccccccc|c}
\toprule
\multirow{2}{*}{\textbf{Methods}} & \multicolumn{12}{c}{\textbf{Edit \& Test Languages}} & \\
& \textbf{en} & \textbf{zh} & \textbf{cz} & \textbf{vi} & \textbf{tr} & \textbf{fr} & \textbf{es} & \textbf{de} & \textbf{ru} & \textbf{du} & \textbf{pt} & \textbf{th} & \textbf{avg}\\
\midrule
\multicolumn{14}{c}{\textbf{Llama3.1-8B, MEMIT}} \\
\midrule
Sum	&0.00 &0.00 &0.00 &0.00 &0.00 &0.00 &0.00 &0.00 &0.00 &0.00 &0.00 &0.00 &0.00\\
Mean &47.66 &46.12 &47.44 &40.85 &43.60 &46.63 &46.14 &50.93 &46.01 &46.88 &48.27 &43.53 &46.17\\
TSVM &62.72 &51.73 &56.72 &50.80 &52.86 &57.12 &56.44 &61.78 &53.40 &56.80 &57.68 &47.24 &55.44\\
Sum-Cov	&\textbf{66.41} &\textbf{55.23} &\textbf{60.01} &\textbf{54.76} &\textbf{56.89} &\textbf{62.03} &\textbf{60.94} &\textbf{65.28} &\textbf{60.37} &\textbf{60.73} &\textbf{60.92} &\textbf{51.83} &\textbf{59.62}\\
Mean-Cov &39.72 &43.28 &39.20 &33.95 &36.91 &38.16 &37.87 &40.84 &41.75 &38.45 &38.90 &41.56 &39.22\\
TSVM-Cov &54.33 &49.38 &51.05 &44.25 &47.81 &51.28 &50.77 &55.67 &50.85 &50.98 &51.76 &45.32 &50.29\\
Mono &70.42 &61.14 &64.63 &60.17 &61.62 &66.17 &65.16 &69.21 &64.45 &63.23 &65.25 &54.75 &63.85\\
\midrule
\multicolumn{14}{c}{\textbf{Llama3.1-8B, AlphaEdit}} \\
\midrule
Sum	&0.00 &0.00 &0.00 &0.00 &0.00 &0.00 &0.00 &0.00 &0.00 &0.00 &0.00 &0.00 &0.00\\
Mean &48.10 &46.62 &47.40 &41.43 &44.04 &46.74 &46.53 &50.97 &46.21 &47.13 &48.07 &43.82 &46.42\\
TSVM &61.08 &50.64 &55.60 &49.68 &51.78 &55.97 &55.06 &60.44 &52.12 &55.33 &55.97 &46.28 &54.16\\
Sum-Cov	&\textbf{65.22} &\textbf{54.39} &\textbf{58.44} &\textbf{53.56} &\textbf{55.36} &\textbf{60.26} &\textbf{59.31} &\textbf{64.01} &\textbf{57.55} &\textbf{59.03} &\textbf{59.09} &\textbf{49.27} &\textbf{57.96}\\
Mean-Cov &39.73 &43.20 &39.16 &34.03 &36.91 &38.26 &37.78 &40.88 &41.68 &38.38 &39.08 &41.64 &39.23\\
TSVM-Cov &52.29 &49.27 &48.16 &43.23 &46.06 &48.99 &48.53 &53.13 &49.02 &48.95 &49.25 &44.68 &48.46\\
Mono &69.52 &59.03 &62.74 &58.49 &59.43 &64.50 &63.71 &67.92 &63.21 &61.42 &63.66 &52.76 &62.20\\
\bottomrule
\end{tabular}%
}
\end{center}
\end{table}

\begin{table}[t]
\caption{Comparison of merging methods on MzsRE dataset with backbone Qwen2.5-7B. Averaged accuracies are reported, unit is percentage. The rank ratio $r$ for TSVM and TSVM-Cov are 0.188, 0.208 for MEMIT, 0.188, 0.396 for AlphaEdit, respectively. Weight scale is set to 1.0.}
\label{tab2}
\begin{center}
\resizebox{\textwidth}{!}{%
\begin{tabular}{c|cccccccccccc|c}
\toprule
\multirow{2}{*}{\textbf{Methods}} & \multicolumn{12}{c}{\textbf{Edit \& Test Languages}} & \\
& \textbf{en} & \textbf{zh} & \textbf{cz} & \textbf{vi} & \textbf{tr} & \textbf{fr} & \textbf{es} & \textbf{de} & \textbf{ru} & \textbf{du} & \textbf{pt} & \textbf{th} & \textbf{avg}\\
\midrule
\multicolumn{14}{c}{\textbf{Qwen2.5-7B, MEMIT}} \\
\midrule
Sum	&0.01 &0.00 &0.05 &0.00 &0.01 &0.00 &0.01 &0.00 &0.00 &0.00 &0.00 &0.02 &0.01\\
Mean &48.94 &53.16 &47.50 &43.06 &44.36 &47.52 &47.01 &49.81 &48.44 &46.38 &48.03 &49.04 &47.77\\
TSVM &63.25 &61.46 &53.48 &48.20 &47.60 &56.47 &57.74 &58.43 &55.87 &54.99 &56.42 &52.38 &55.52\\
Sum-Cov	&\textbf{66.90} &\textbf{66.69} &\textbf{59.77} &\textbf{53.60} &\textbf{53.58} &\textbf{60.88} &\textbf{62.94} &\textbf{62.59} &\textbf{63.61} &\textbf{59.84} &\textbf{61.95} &\textbf{59.66} &\textbf{61.00}\\
Mean-Cov &40.38 &49.45 &40.87 &37.19 &39.10 &40.45 &39.31 &40.51 &45.17 &38.49 &40.69 &47.38 &41.58\\
TSVM-Cov &62.73 &61.67 &53.08 &49.18 &47.32 &55.91 &57.33 &57.68 &55.89 &53.64 &56.08 &52.81 &55.28\\
Mono &71.92 &72.10 &65.85 &61.95 &63.37 &68.25 &69.57 &67.24 &67.11 &65.85 &67.95 &63.96 &67.09\\
\midrule
\multicolumn{14}{c}{\textbf{Qwen2.5-7B, AlphaEdit}} \\
\midrule
Sum	&0.00 &0.01 &0.00 &0.00 &0.00 &0.01 &0.00 &0.00 &0.02 &0.00 &0.00 &0.02 &0.00\\
Mean &52.18 &53.82 &49.45 &44.46 &45.31 &49.50 &49.07 &52.37 &49.33 &48.88 &49.93 &49.17 &49.46\\
TSVM &65.07 &\textbf{64.13} &\textbf{57.10} &\textbf{51.26} &\textbf{51.12} &\textbf{58.16} &\textbf{59.92} &\textbf{60.58} &\textbf{59.05} &\textbf{57.64} &\textbf{59.04} &\textbf{55.01} &\textbf{58.17}\\
Sum-Cov	&63.07 &59.50 &50.53 &46.68 &39.93 &56.00 &58.96 &55.05 &54.92 &52.09 &55.94 &51.13 &53.65\\
Mean-Cov &41.05 &50.05 &41.36 &37.48 &39.59 &41.09 &39.77 &40.97 &45.43 &38.83 &41.16 &47.54 &42.03\\
TSVM-Cov &65.36 &62.69 &55.49 &50.62 &49.73 &58.92 &59.98 &59.85 &57.04 &56.56 &59.41 &53.43 &57.42\\
Mono &72.29 &66.29 &68.37 &64.16 &65.40 &70.46 &71.32 &68.56 &69.75 &67.88 &69.60 &68.16 &68.52\\
\bottomrule
\end{tabular}%
}
\end{center}
\end{table}

\subsubsection{Effect of Weight Scale (\textbf{RQ3})}
\begin{figure}[t]
\centering
\subfloat[]{\includegraphics[width=0.45\linewidth]{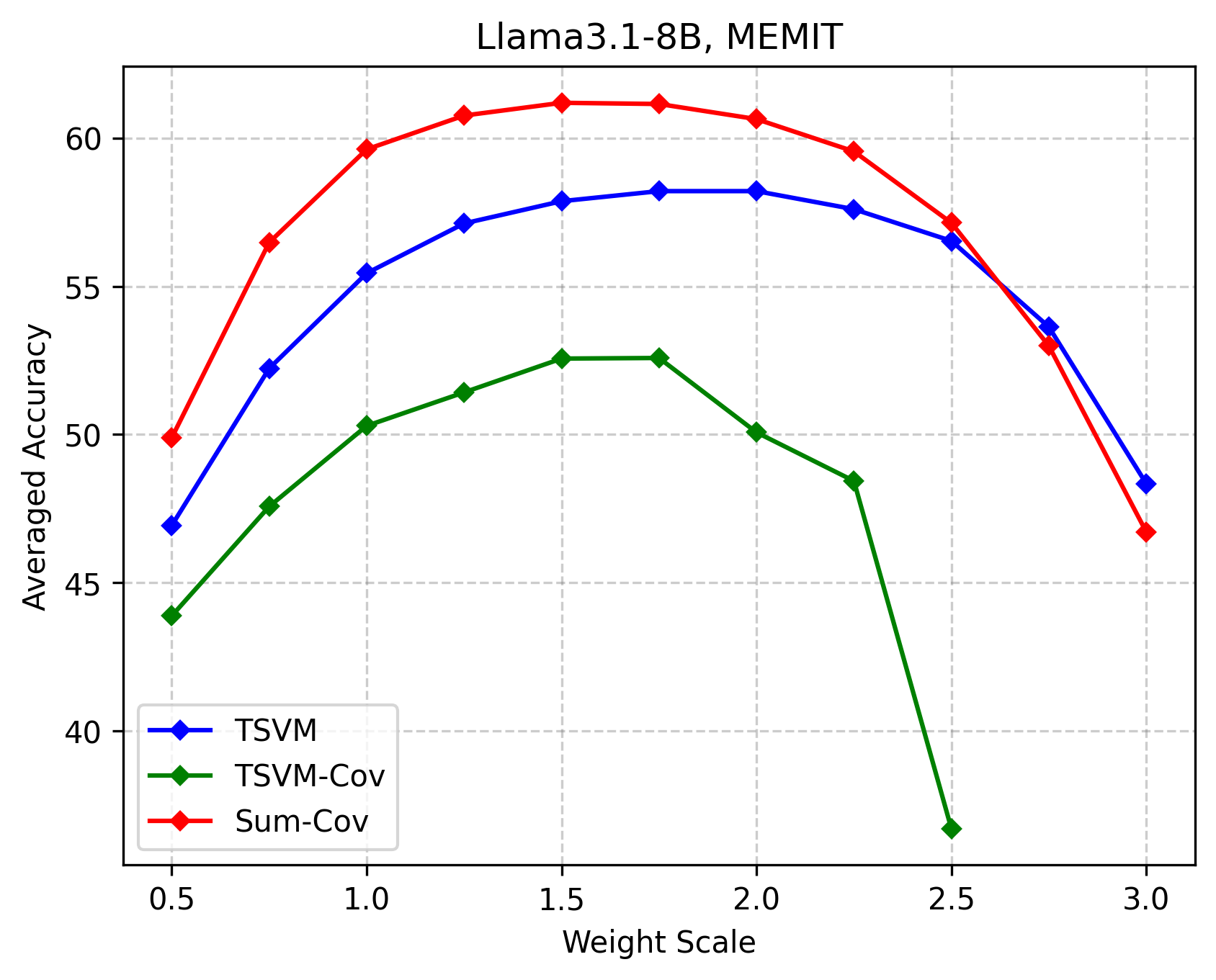}}
\subfloat[]{\includegraphics[width=0.45\linewidth]{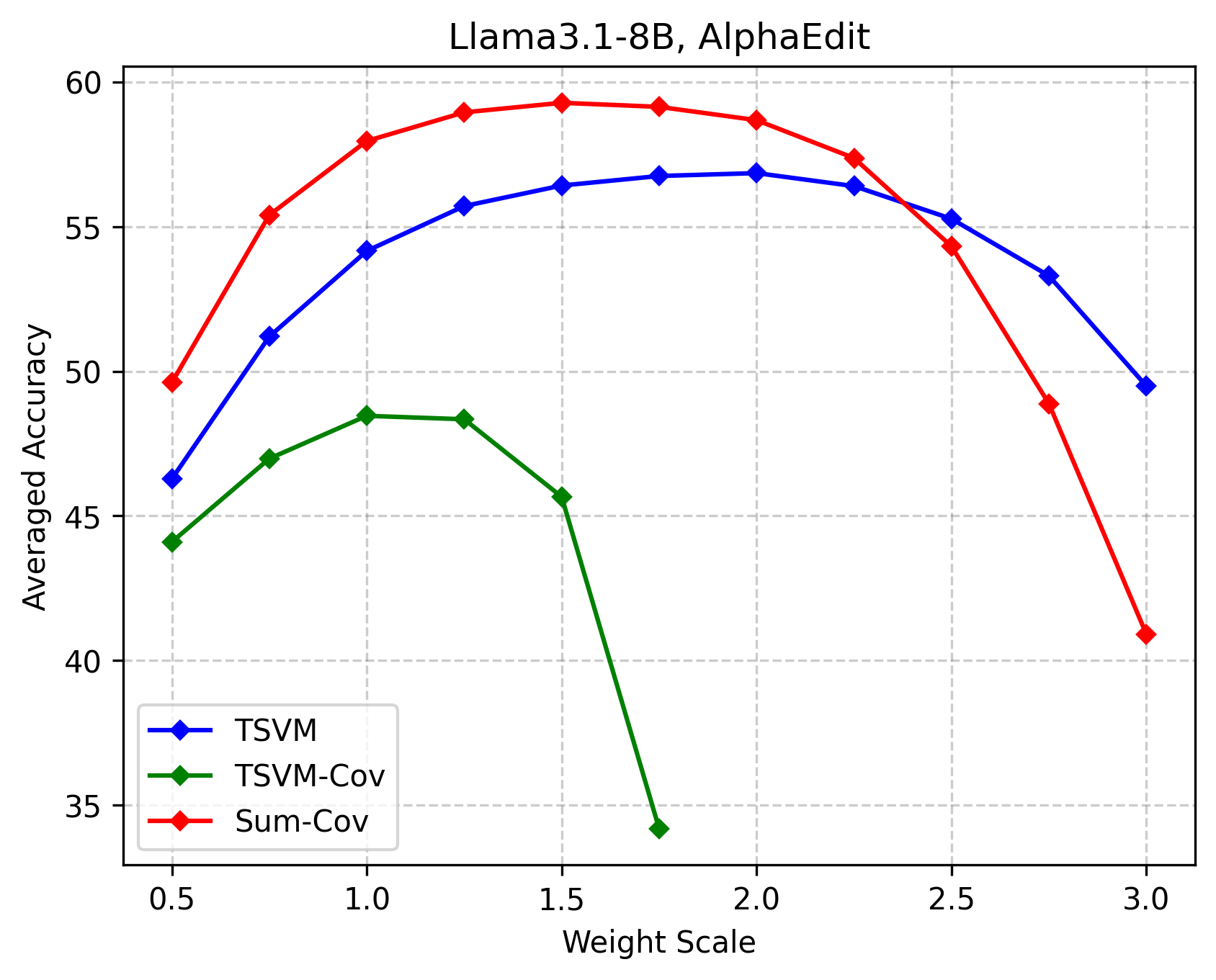}}\\
\subfloat[]{\includegraphics[width=0.45\linewidth]{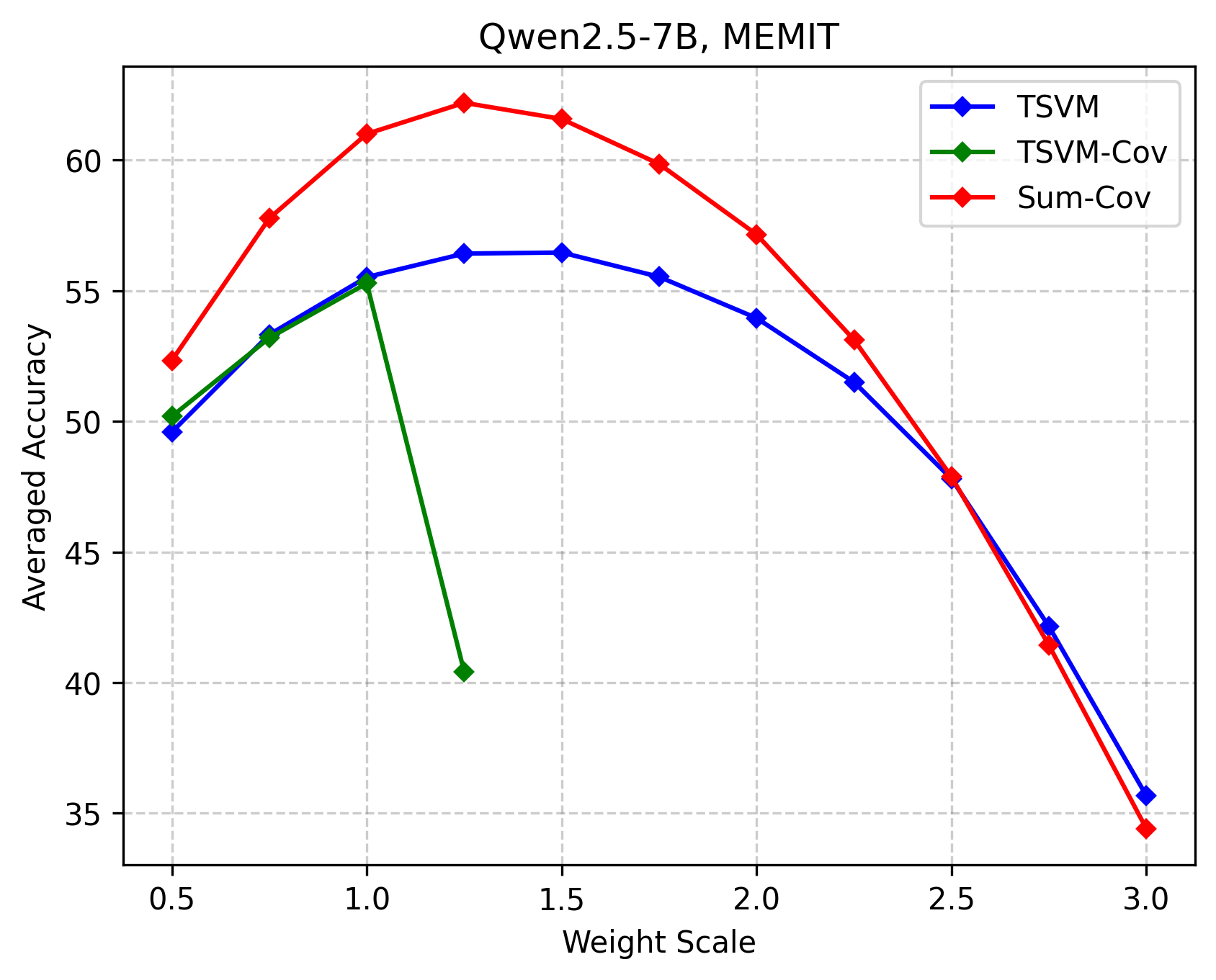}}
\subfloat[]{\includegraphics[width=0.45\linewidth]{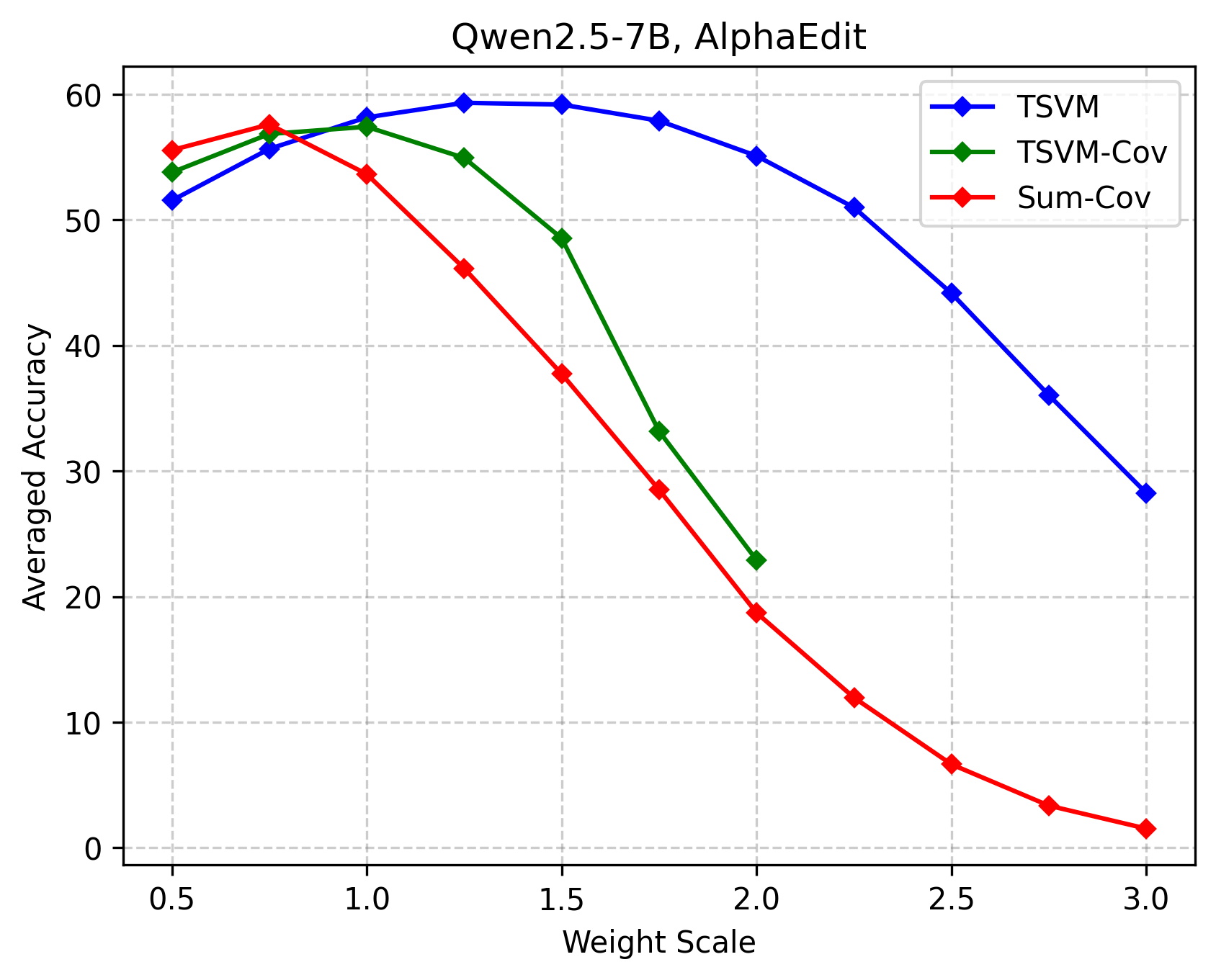}}
\caption{Effect of scaling factor $\alpha$ on TSVM, TSVM-Cov, and Sum-Cov. Accuracies are averaged across all languages.}
\label{fig1}
\end{figure}

We also investigate the effect of weight scaling factor $\alpha$ in the \eqref{eq:update}, which is ignored in previously published work in the field of locate-then-edit framework. To the best of our knowledge, this study represents the first investigation of the effect of scaling factor in the MKE. The results are shown in Figure~\ref{fig1} and we have following key observation:

\begin{itemize}
    \item \textbf{Key Observation: Properly scaled weights achieve higher performances.} All curves are convex and the optimal positions are marginally higher than 1.0 (which is the default weight scale) in most cases.
\end{itemize}

\subsubsection{Effect of Rank Ratio $r$ (\textbf{RQ3})}
\begin{figure}[t]
\centering
\subfloat[]{\includegraphics[width=0.45\linewidth]{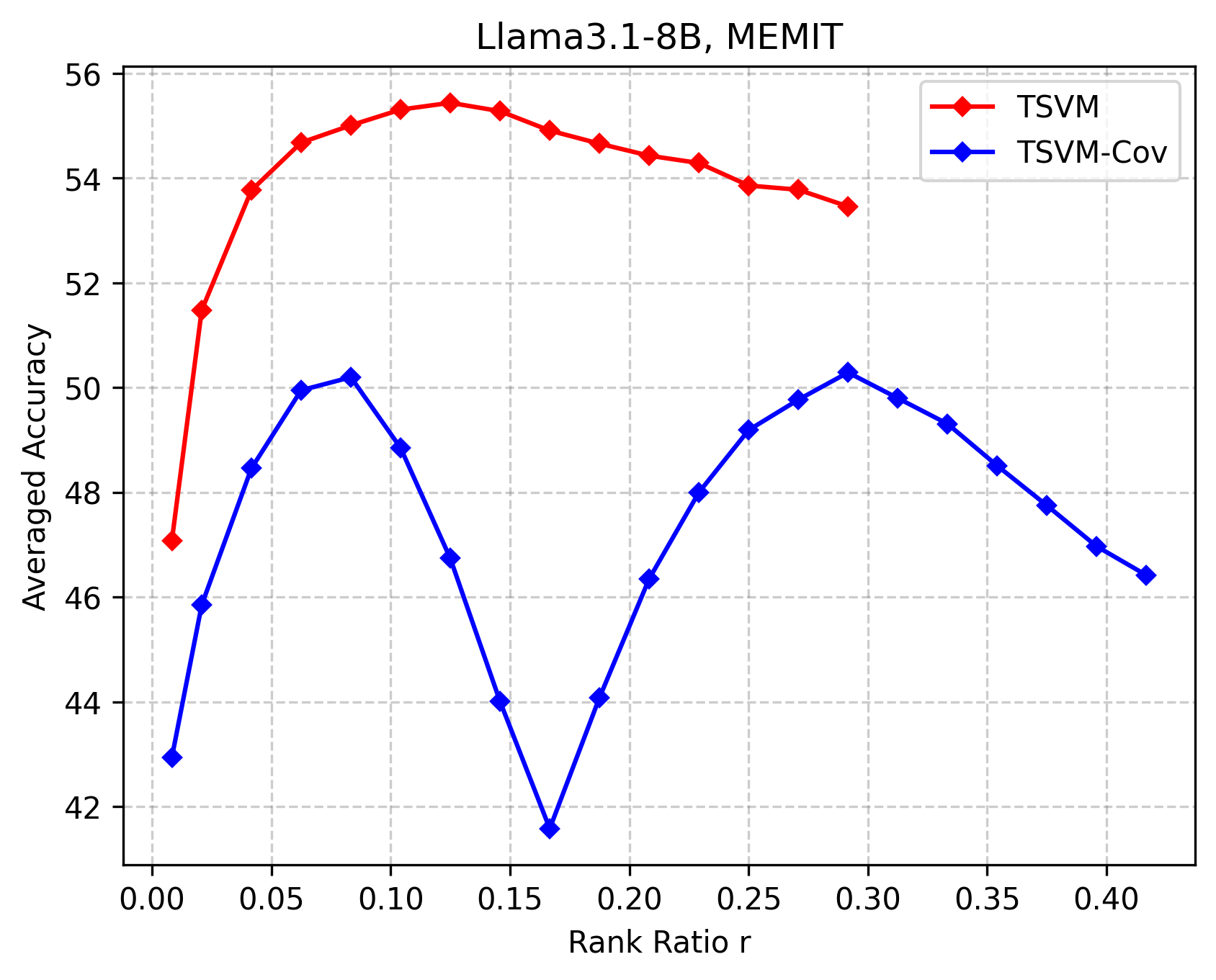}}
\subfloat[]{\includegraphics[width=0.45\linewidth]{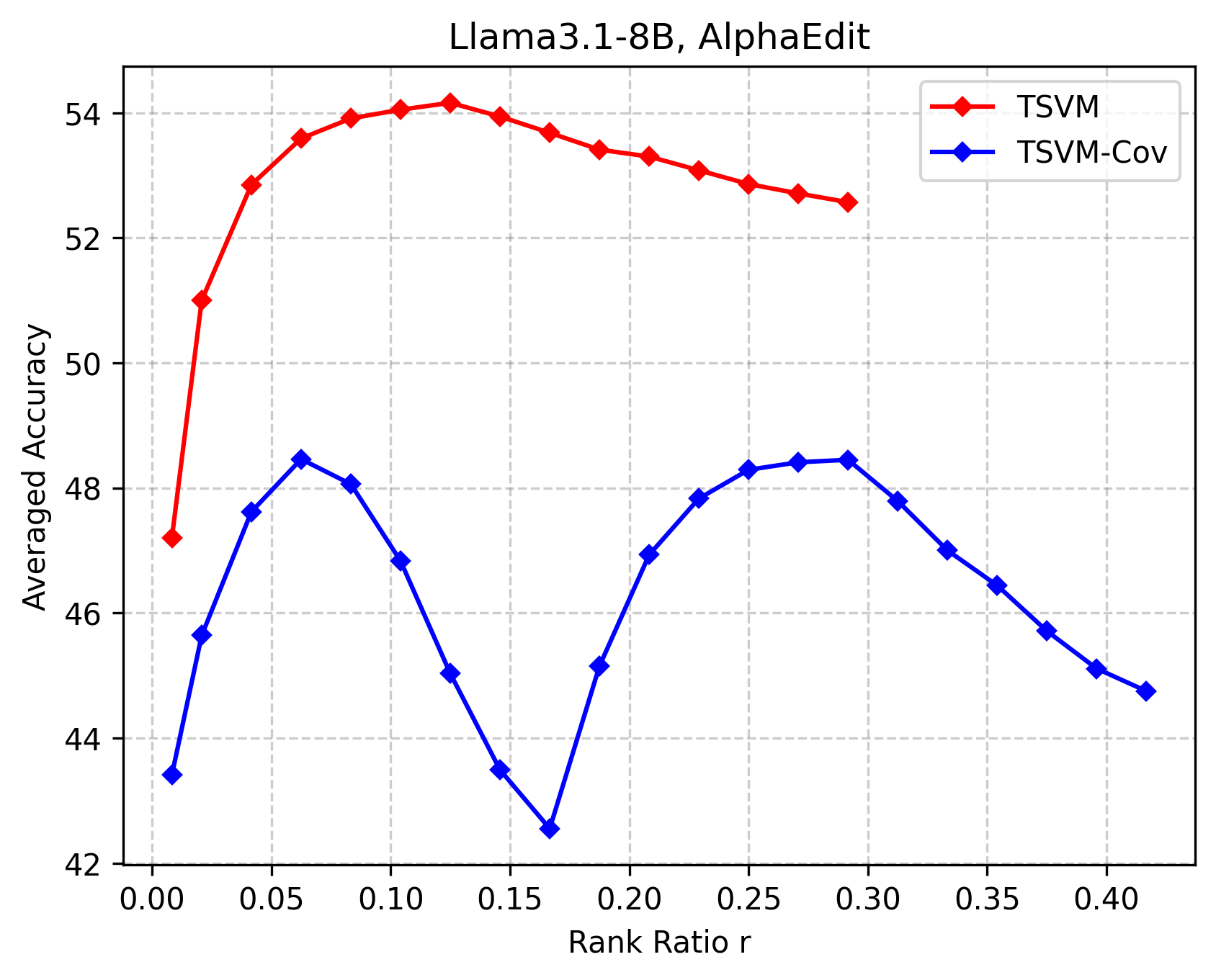}}\\
\subfloat[]{\includegraphics[width=0.45\linewidth]{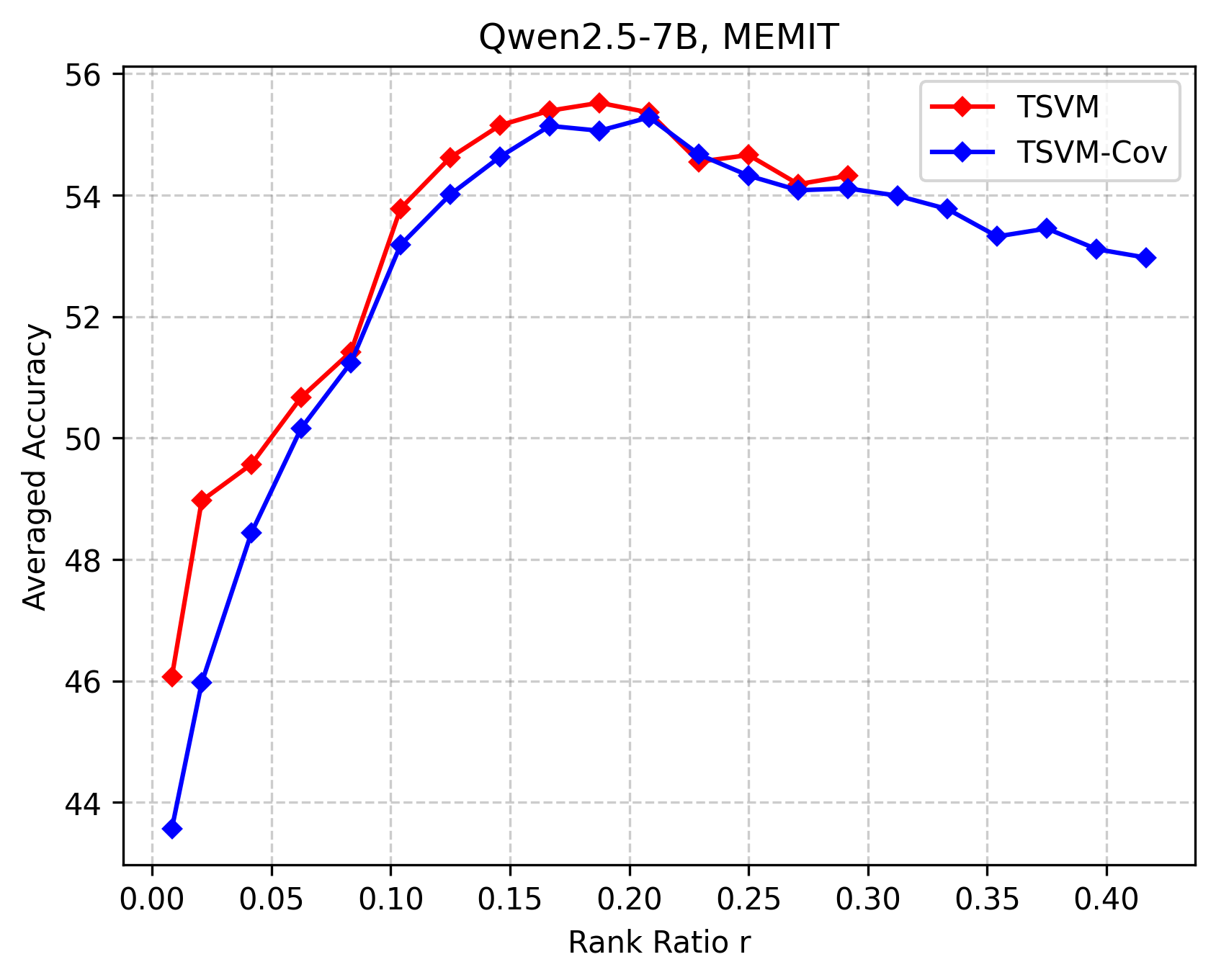}}
\subfloat[]{\includegraphics[width=0.45\linewidth]{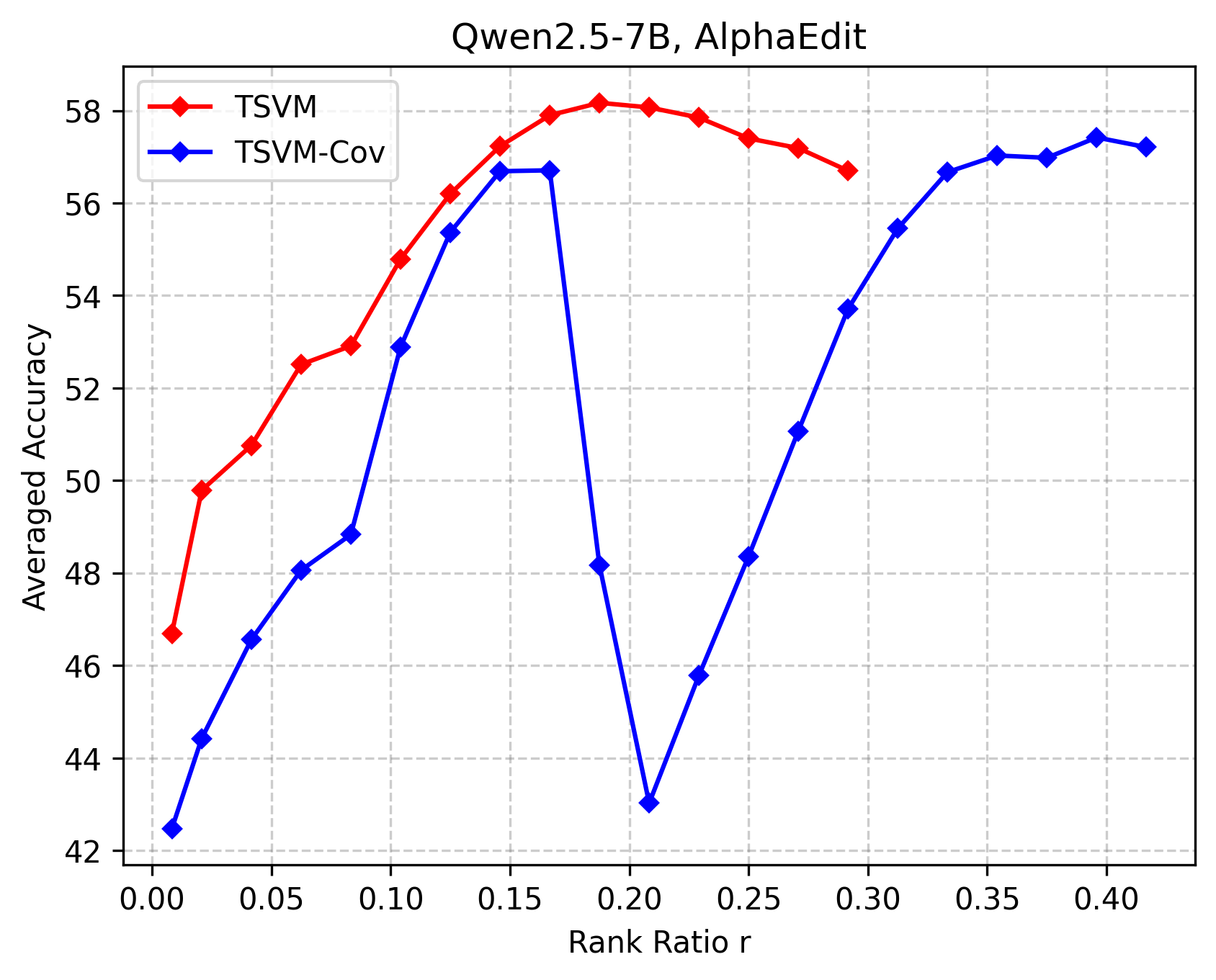}}
\caption{Effect of rank ratio $r$ on TSVM, TSVM-Cov. Accuracies are averaged across all languages.}
\label{fig2}
\end{figure}

The rank ratio $r$ in the \eqref{eq:factor-r} determines the compression ratio of TSVM based methods. In the original work of TSVM \citep{tsv}, they use a fixed ratio (0.083 in our case). Here we investigate the effect of rank ratio $r$ and the results are shown in Figure~\ref{fig2}. We have following observations:
\begin{itemize}
    \item \textbf{Observation 1: In most cases, the curve of TSVM has one modality while TSVM-Cov has at least two.} For some unknown reasons, performance of TSVM-Cov drops dramatically during some range of $r$ between 0.15 to 0.20, and then increases. However, TSVM maintains a single convex curve through our evaluated range.
    \item \textbf{Observation 2: Properly low rank ratio tends to achieve optimal performance.} It's not surprising because of the low-rank nature of knowledge editing vectors \citep{rome}. In many work on model merging \citep{tsv}, they report rank reduction can boost merging performance. In the MKE, this result is the first analysis of the relation between the rank of $\bm{\Delta}$ and the performance.
\end{itemize}

\section{Discussion}
\subsection{Effect of Vector Merging Framework (\textbf{RQ1})}
Our results show that the choice of merging framework has a substantial effect on MKE performance. Among the evaluated methods, Sum-Cov consistently provides the strongest overall results across most backbone and base-method combinations. This finding suggests that sharing covariance across languages is more important than applying a sophisticated merging rule after language-specific updates have already diverged. In contrast, naive summation without shared covariance almost completely fails, indicating that multilingual editing vectors are not directly compatible when computed independently.

Another important observation is that averaging-based methods do not offer a reliable compromise. Although Mean improves dramatically over Sum, it still remains clearly below Sum-Cov, and Mean-Cov performs even worse in most settings. This pattern indicates that simply reducing the magnitude of updates cannot resolve multilingual interference by itself. Instead, the interaction among language-specific key distributions appears to be a more fundamental source of the problem.

More broadly, the results show that current vector merging methods do not eliminate the gap between multilingual and monolingual editing. Even the best multilingual setting remains noticeably below the monolingual upper bound in all four experimental configurations. Therefore, the main challenge in MKE is not merely how to merge language-specific updates, but how to construct updates that are mutually compatible across languages before or during merging.

\subsection{Effect of TSVM (\textbf{RQ2})}
TSVM was motivated by prior evidence that low-rank task representations can reduce interference in model merging. In our setting, TSVM is clearly more effective than simple Sum and often competitive with strong baselines, which suggests that low-rank structure is indeed relevant for multilingual editing vectors. However, TSVM does not consistently outperform Sum-Cov, and TSVM-Cov improves over TSVM only in limited cases. In other words, rank-aware merging alone is insufficient to solve multilingual interference.

One notable exception appears in the Qwen + AlphaEdit setting, where TSVM and TSVM-Cov both outperform Sum-Cov. This result suggests that the usefulness of TSVM depends on the geometry of the underlying editing vectors, which in turn may depend on both the backbone model and the base KE algorithm. However, because this behavior is not reproduced in the other settings, we interpret it as evidence of conditional effectiveness rather than a generally superior strategy.

Taken together, these findings answer RQ2 conservatively: TSVM can mitigate multilingual interference to some extent, but only under limited conditions. Future work should examine when TSVM is beneficial by analyzing the spectral structure of editing vectors and the degree of alignment between languages.

\subsection{Effect of Weight Scaling Factor (\textbf{RQ3})}
The effect of weight scaling is one of the most practically important findings of this study. In most settings, the best performance is obtained not at the default scale of $1.0$, but at a slightly larger value. This result shows that the magnitude of the closed-form update is not automatically well calibrated, even when the editing direction itself is useful. From an optimization perspective, the scaling factor plays a role analogous to a learning rate, although locate-then-edit methods are derived analytically rather than through iterative training.

One possible explanation is that the scale of the computed update is sensitive to several handcrafted components in the closed-form solutions of \eqref{eq:memit} and \eqref{eq:alpha-eidt}. In particular, both the covariance-related term and the estimated target-value term $\mV_{req}$ can vary substantially in magnitude. As a result, two editing methods with similar directional quality may still differ in performance because one produces an under-scaled or over-scaled update.

This finding also has practical implications. If scaling is tuned appropriately, practitioners may obtain a noticeable gain without changing the editing algorithm itself. At the same time, exhaustive search over the scaling factor is expensive. A useful direction for future work is therefore to develop principled or data-driven rules for choosing the scaling factor automatically.

\subsection{Effect of Rank Compression Ratio (\textbf{RQ3})}
The rank compression ratio controls how much information from each language-specific editing vector is retained during TSVM. Our results show that relatively low rank often leads to the best performance, which is consistent with the view that the most important multilingual editing signals are concentrated in a compact subspace.

At the same time, the behavior of TSVM-Cov is more complex than that of TSVM. In several settings, TSVM-Cov exhibits a non-smooth performance curve with abrupt drops in a narrow range of rank ratios. This instability suggests that shared covariance and low-rank reconstruction interact in a nontrivial way.

\subsection{Limitations of This Work}
This study has several limitations. First, we evaluate only two backbone models and two locate-then-edit base methods, so our conclusions may not transfer uniformly to other architectures or editing paradigms. Second, our experiments focus on one-step batch editing on MzsRE, which means the results do not directly address sequential editing, continual editing, or other multilingual editing benchmarks. In addition, our search over weight scales and rank ratios is empirical rather than theoretically grounded. Although this is sufficient to reveal important trends, it does not yet provide a predictive framework for selecting hyperparameters.

\section{Conclusions}
In this paper, we investigated whether vector merging methods can improve multilingual knowledge editing for large language models. We evaluated six merging variants across two backbone models and two locate-then-edit base methods under a large-scale multilingual batch-editing setting covering 12 languages. Our results show that shared-covariance summation is the most reliable overall strategy, while TSVM can be beneficial only in limited settings.

More importantly, the study highlights two broader findings. First, multilingual interference remains a major bottleneck: none of the evaluated merging methods closes the gap between multilingual and monolingual editing. Second, the scaling factor and rank compression ratio materially affect performance, showing that the quality of a closed-form edit depends not only on direction but also on magnitude and spectral structure.

These findings suggest that future progress in MKE will likely require methods that model cross-lingual compatibility more explicitly, rather than relying only on post hoc merging of independently computed updates. We hope this work provides a strong empirical basis for that direction and serves as a useful benchmark for future research on multilingual locate-then-edit methods.

\subsubsection*{Acknowledgments}
This research was partly supported by Basic Science Research Program through the National Research Foundation of Korea (NRF) funded by the Ministry of Education (RS-2022-NR070870), by Institute of Information \& communications Technology Planning \& Evaluation (IITP) grant funded by the Korea government (MSIT) (No.RS-2019-II191906, Artificial Intelligence Graduate School Program at POSTECH), and by the Gyeongsangbuk-do RISE (Regional Innovation System \& Education) project (B0080527002599). We thank technical and financial support that provided by Designovel Co., Ltd.


\begin{thebibliography}{99}

\bibitem[OpenAI(2023)]{gpt}
OpenAI. GPT-4 Technical Report. Technical Report, 2023.

\bibitem[Grattafiori(2024)]{llama}
Llama Team, Meta AI. The Llama 3 Herd of Models. Technical Report, 2024.

\bibitem[Yang(2024)]{qwen}
Qwen Team, Alibaba Group. Qwen2 Technical Report. Technical Report, 2024.

\bibitem[Google(2025)]{gemini}
Gemini Team, Google. Gemini 2.5: Pushing the Frontier with Advanced Reasoning, Multimodality, Long Context, and Next Generation Agentic Capabilities. Technical Report, 2025.

\bibitem[Vaswani et~al.(2017)]{attn}
Vaswani, A.; Shazeer, N.; Parmar, N.; Uszkoreit, J.; Jones, L.; Gomez, A.N.; Kaiser, {\L}.; Polosukhin, I. Attention Is All You Need. In Proceedings of the Annual Conference on Neural Information Processing Systems (NIPS), 2017.

\bibitem[Hu et~al.(2022)]{lora}
Hu, E.J.; Shen, Y.; Wallis, P.; Allen-Zhu, Z.; Li, Y.; Wang, S.; Wang, L.; Chen, W. LoRA: Low-Rank Adaptation of Large Language Models. In Proceedings of the International Conference on Learning Representations, 2022.

\bibitem[Mitchell et~al.(2022a)]{mend}
Mitchell, E.; Lin, C.; Bosselut, A.; Finn, C.; Manning, C.D. Fast Model Editing at Scale. In Proceedings of the International Conference on Learning Representations, 2022.

\bibitem[Yao et~al.(2023)]{survey1}
Yao, Y.; Wang, P.; Tian, B.; Cheng, S.; Li, Z.; Deng, S.; Chen, H.; Zhang, N. Editing large language models: Problems, methods, and opportunities. In Proceedings of the 2023 Conference on Empirical Methods in Natural Language Processing, 2023.

\bibitem[Meng et~al.(2022)]{rome}
Meng, K.; Bau, D.; Andonian, A.; Belinkov, Y. Locating and Editing Factual Associations in GPT. In Advances in Neural Information Processing Systems, 2022.

\bibitem[Meng et~al.(2023)]{memit}
Meng, K.; Sharma, A.S.; Andonian, A.; Belinkov, Y.; Bau, D. Mass-Editing Memory in a Transformer. In Proceedings of the International Conference on Learning Representations, 2023.

\bibitem[Fang et~al.(2025)]{alpha-edit}
Fang, J.; Jiang, H.; Wang, K.; Ma, Y.; Shi, J.; Wang, X.; He, X.; Chua, T.-S. AlphaEdit: Null-Space Constrained Knowledge Editing for Language Models. In Proceedings of the International Conference on Learning Representations, 2025.

\bibitem[Wang et~al.(2024)]{cross}
Wang, J.; Liang, Y.; Sun, Z.; Cao, Y.; Xu, J.; Meng, F. Cross-Lingual Knowledge Editing in Large Language Models. In Proceedings of the 62nd Annual Meeting of the Association for Computational Linguistics, 2024.

\bibitem[Zhang et~al.(2025)]{lafn}
Zhang, X.; Liang, Y.; Meng, F.; Zhang, S.; Chen, Y.; Xu, J.; Zhou, J. Multilingual Knowledge Editing with Language-Agnostic Factual Neurons. In Proceedings of the 31st International Conference on Computational Linguistics, 2025.

\bibitem[Gargiulo et~al.(2025)]{tsv}
Gargiulo, A.A.; Crisostomi, D.; Bucarelli, M.S.; Scardapane, S.; Silvestri, F.; Rodol{\`a}, E. Task Singular Vectors: Reducing Task Interference in Model Merging. In Proceedings of the IEEE/CVF Conference on Computer Vision and Pattern Recognition, 2025.

\bibitem[Wang et~al.(2023)]{mzsre}
Wang, W.; Haddow, B.; Birch, A. Retrieval-augmented Multilingual Knowledge Editing. Preprint, arXiv:2312.13040, 2023.

\bibitem[Zheng et~al.(2023)]{ike}
Zheng, C.; Li, L.; Dong, Q.; Fan, Y.; Wu, Z.; Xu, J.; Chang, B. Can We Edit Factual Knowledge by In-Context Learning? In Proceedings of the 2023 Conference on Empirical Methods in Natural Language Processing, 2023.

\bibitem[De~Cao et~al.(2021)]{knowledge-editor}
De~Cao, N.; Aziz, W.; Titov, I. Editing Factual Knowledge in Language Models. In Proceedings of the 2021 Conference on Empirical Methods in Natural Language Processing, 2021, pp. 6491--6506.

\bibitem[Mitchell et~al.(2022b)]{serac}
Mitchell, E.; Lin, C.; Bosselut, A.; Manning, C.D.; Finn, C. Memory-Based Model Editing at Scale. In Proceedings of the 39th International Conference on Machine Learning, 2022, pp. 15817--15831.

\bibitem[Dai et~al.(2022)]{knowledge-neurons}
Dai, D.; Dong, L.; Hao, Y.; Sui, Z.; Chang, B.; Wei, F. Knowledge Neurons in Pretrained Transformers. In Proceedings of the 60th Annual Meeting of the Association for Computational Linguistics, 2022, pp. 8493--8502.

\bibitem[Khandelwal et~al.(2024)]{crolin-mquake}
Khandelwal, A.; Singh, H.; Gu, H.; Chen, T.; Zhou, K. Cross-Lingual Multi-Hop Knowledge Editing. In Findings of the Association for Computational Linguistics: EMNLP 2024, 2024, pp. 11995--12015.

\bibitem[Li et~al.(2024)]{pmet}
Li, X.; Li, S.; Song, S.; Yang, J.; Ma, J.; Yu, J. PMET: Precise Model Editing in a Transformer. In Proceedings of the 38th Annual AAAI Conference on Artificial Intelligence, 2024.

\bibitem[Wortsman et~al.(2022)]{model-soups}
Wortsman, M.; Ilharco, G.; Gadre, S.Y.; Roelofs, R.; Gontijo Lopes, R.; Morcos, A.; Namkoong, H.; Farhadi, A.; Carmon, Y.; Kornblith, S.; Schmidt, L. Model Soups: Averaging Weights of Multiple Fine-Tuned Models Improves Accuracy without Increasing Inference Time. In Proceedings of the 39th International Conference on Machine Learning, 2022.

\bibitem[Ilharco et~al.(2023)]{task-arithmetic}
Ilharco, G.; Ribeiro, M.T.; Wortsman, M.; Schmidt, L.; Hajishirzi, H.; Farhadi, A. Editing Models with Task Arithmetic. In Proceedings of the International Conference on Learning Representations, 2023.

\bibitem[Yadav et~al.(2023)]{ties-merging}
Yadav, P.; Tam, D.; Choshen, L.; Raffel, C.; Bansal, M. TIES-Merging: Resolving Interference When Merging Models. In Advances in Neural Information Processing Systems, 2023.

\bibitem[Geva et~al.(2021)]{key-value}
Geva, M.; Schuster, R.; Berant, J.; Levy, O. Transformer Feed-Forward Layers Are Key-Value Memories. In Proceedings of the 2021 Conference on Empirical Methods in Natural Language Processing, 2021.

\bibitem[Lang(2012)]{linalg}
Lang, S. Introduction to Linear Algebra. Springer Science \& Business Media, 2012.

\bibitem[Paszke et~al.(2019)]{pytorch}
Paszke, A.; et~al. PyTorch: An Imperative Style, High-Performance Deep Learning Library. In Advances in Neural Information Processing Systems, 2019.

\end{thebibliography}
\end{document}